\documentclass[conference]{IEEEtran}
\usepackage{cite}
\usepackage{amsmath,amssymb,amsfonts}
\usepackage{algorithmic}
\usepackage{graphicx}
\usepackage{textcomp}
\usepackage{xcolor}
\usepackage{url}
\usepackage[hidelinks]{hyperref}
\usepackage{array}

\begin{document}
\pagestyle{plain}

\title{Handling imbalance and few-sample size in ML based Onion disease classification}

\author{
    \IEEEauthorblockN{
        Abhijeet Manoj Pal\IEEEauthorrefmark{1},
        Rajbabu Velmurugan\IEEEauthorrefmark{2}
    }
    \IEEEauthorblockA{\IEEEauthorrefmark{1}Centre for Machine Intelligence and Data Science, IIT Bombay, Mumbai, India \\
    Email: 200100107@iitb.ac.in}
    \\
    \IEEEauthorblockA{\IEEEauthorrefmark{2}Department of Electrical Engineering, IIT Bombay, Mumbai, India\\
    Email: rajbabu@ee.iitb.ac.in}
}

\maketitle

\begin{abstract}
Accurate classification of pests and diseases plays a vital role in precision agriculture, enabling efficient identification, targeted interventions, and preventing their further spread. However, current methods primarily focus on binary classification, which limits their practical applications, especially in scenarios where accurately identifying the specific type of disease or pest is essential. We propose a robust deep learning based model for multi-class classification of onion crop diseases and pests. We enhance a pre-trained Convolutional Neural Network (CNN) model by integrating attention based modules and employing comprehensive data augmentation pipeline to mitigate class imbalance. We propose a model which gives 96.90\% overall accuracy and 0.96 F1 score on real-world field image dataset. This model gives better results than other approaches using the same datasets.
\end{abstract}


\section{Introduction}
India is the largest onion producer in the world, accounting for approximately 28.66\% of the total production in the world \cite{faostat_onion}. Onion pests and diseases lead to significant crop losses in India. Onion has seven main types of pests and diseases. By identifying pests and diseases accurately, targeted treatments can be implemented to treat and prevent further spread of diseases and pests.
Different onion diseases and pests need separate focused treatments, hence it is extremely necessary to classify onion diseases and pests separately rather than classification as healthy or non-healthy. 

Previous work such as Gbadebo et al. \cite{gbadebo2022} focused on binary classification of onion leaves into healthy and unhealthy categories. The data collected were from internet-sites and a handful from actual farm. However, this level of classification is limited for practical agricultural use, where identifying the exact type of disease or pest is crucial for targeted treatment. In this work, we address this gap by classifying seven specific onion diseases and pests and healthy crop using deep learning techniques.

Raj et al. \cite{raj2025} proposed an improved model based on YOLOv8s for the detection of onion foliar disease. It does identify specific diseased regions before classification. However, this adds an additional complexity and overhead of bounding-box before classification. Moreover, it requires region-level annotations. It is a two-step pipleline involving localization and classification, and is susceptible to errors due to annotation inaccuracies, which is exacerbated due to the use of approximately 279 images per class for training. Also, they focus on only five image classes. In image augmentation. Our approach of using Convolutional Neural Network based models work on full leaf images eliminating the need of such annotations and bounding box overhead in inference time. We also use dataset balancing techniques to mitigate the imbalance among the eight classes. 

To address these gaps, this paper compares different CNN based architectures, loss functions, and data imbalance techniques to develop a robust onion pest and disease classification model.

\section{Related Work}
Very little research work was found to be in the area of Onion leaf disease and pest classification. Gbadebo et al. \cite{gbadebo2022} have used healthy and non-healthy classification on images mostly obtained from internet-sites and handful from farms. Raj et al. \cite{raj2025} have used Yolo-v8 model for classification, it increases overhead and due to the less number of images approximately 279 images per class in training for fine-tuning, the model is prone to not classify the data properly, which is observed in the class-wise accuracies not exceeding 0.784. Lokhande et. al. \cite{lokhande2024} conducted a comparative analysis of AlexNet and ResNet-50 models for soyabean and maize crops disease on a dataset which is a combination of Plant Village Dataset and images from agricultural plots and has observed that ResNet-50 model gives better classification results compared to AlexNet model in both cases.  For soybean disease classification, ResNet-50 achieved an accuracy of 97.41\%, compared to 96.40\% with AlexNet. Similarly, for maize disease classification, ResNet-50 achieved 96.74\% accuracy, while AlexNet reached 95.99\%. Zhang et. al. \cite{zhang2018maize} used CNN models for classification of nine types of maize leaves, obtaining high accuracies of 98.9\% using GoogLeNet model. This motivates us to use Convolutional Neural Networks (CNNs) for the classification of Onion Disease and Pest Images.

\section{Materials and Methodology}
\subsection{Dataset}

\begin{figure}[htbp]
    \centering
    \includegraphics[width=0.9\linewidth]{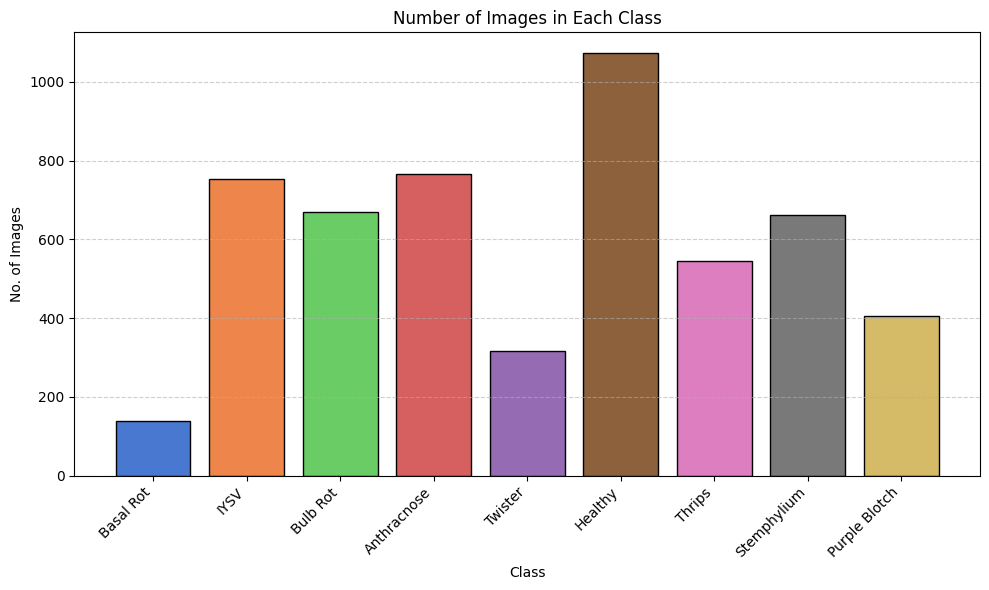}
    \caption{Number of images in each class in the dataset}
    \label{fig:class_distribution}
\end{figure}

\begin{figure}[htbp]
    \centering
    \includegraphics[width=0.9\linewidth]{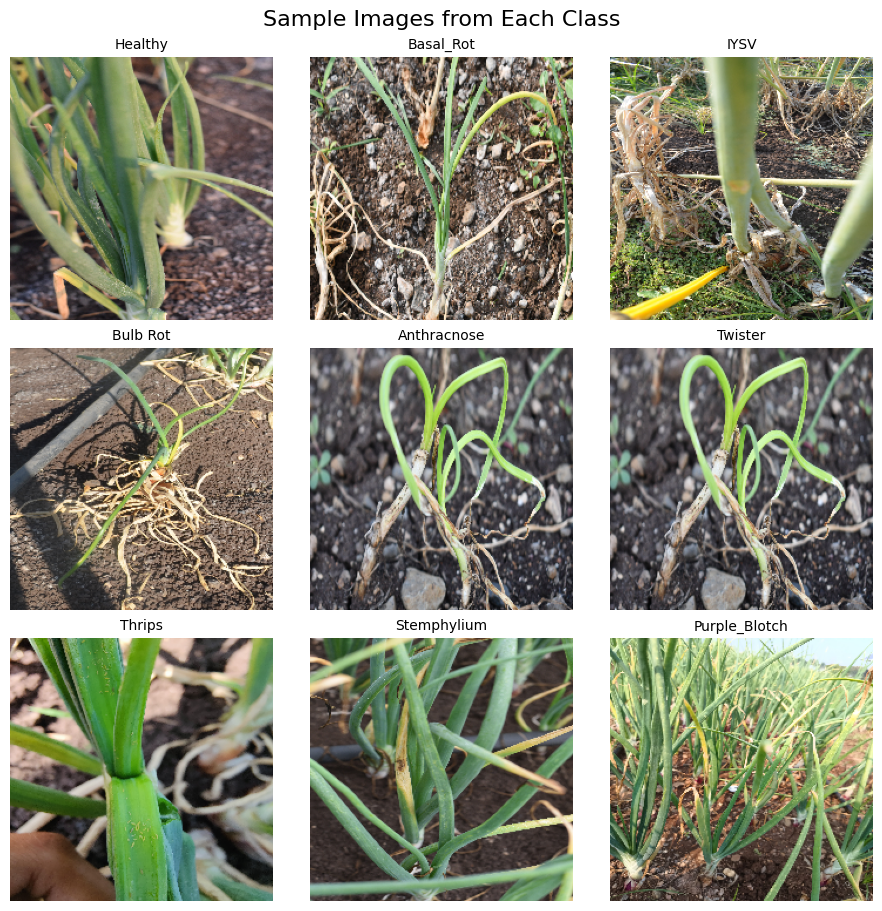}
    \caption{Sammple Images from Each Class}
    \label{fig:dataset_imgs}
\end{figure}

The dataset used is obtained from the Indian Council of Agricultural Research (ICAR)-Directorate of Onion and Garlic Research, Pune, through the TIH Foundation for Technology Innovation Hub, IIT Bombay, Internet of Things and Internet of Everything (IoT \& IoE), Mumbai. The images consist of both those taken via a DSLR camera and a smartphone. The dataset consists of nine classes, Healthy, Basal Rot, IYSV (Iris Yellow Spot Virus), Bulb Rot, Anthracnose, Twister, Thrips, Stemphylium, Purple Blotch. The dataset is imbalanced, as shown in Figure~\ref{fig:class_distribution}. The total number of images are 5330, with maximum for Healthy class, 1072 and minimum for Basal Rot, 140 images. Thus imbalance ratio of 7.6 : 1. A few images of the dataset are as shown in Figure~\ref{fig:dataset_imgs}.

\subsection{Data Pre-Processing and Augmentation}
We employed different data pre-processing pipelines tailored for each model setup:

\subsubsection{Pipeline A: Normal Augmentation}
This pipeline includes resizing images to 224×224, normalization, and basic augmentations such as horizontal flip and rotation.

\subsubsection{Pipeline B : Imbalanced Sampler}
\begin{figure}[htbp]
    \centering
    \includegraphics[width=0.9\linewidth]{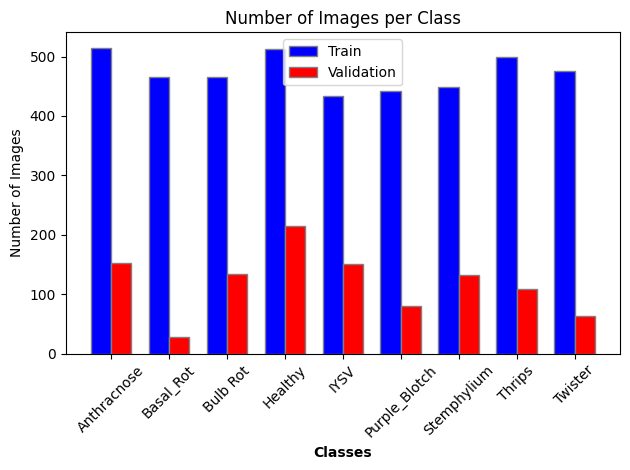}
    \caption{Dataset Distribution after Imbalanced Dataset Sampler}
    \label{fig:imbalanced_sampler}
\end{figure}

As the classes in the dataset are highly imbalanced, an imbalanced dataset sampler was used as proposed by Yang et. al. \cite{yang2021imbalanced}. Here, the majority class are undersampled and the minority class are oversampled. After using the Imbalanced Dataset Sampler the class wise distribution obtained is as shown in Figure~\ref{fig:imbalanced_sampler}

\subsubsection{Pipeline C : Albumentations-Based Augmentation}
The image was randomly subjected to one of the following augmentations, as proposed by Buslaev et al. \cite{buslaev2020albumentations}. A combination of general transformations and class-specific augmentations based on the type of disease were applied. For each image, one of the following augmentations was randomly applied such as, horizontal flip, vertical flip, motion blur, median blur, guassian blur, Hue variations, grid shuffle and coarse dropout. The images were all resized to 224×224. The augmentations were a combination of these. Each transformation was applied with a specific probability to enhance the diversity of the dataset and improve model generalization.

\subsubsection{Pipeline D : Cut-Mix Augmentation}
CutMix was used to enhance the model's generalization as proposed by \cite{yun2019cutmix}. It involves combining two images and adjusting the labels based on the overlapping region. It generates a new training sample by combining two input samples. Let \( x_A \in \mathbb{R}^{W \times H \times C} \) and \( x_B \in \mathbb{R}^{W \times H \times C} \) be two training images, with corresponding labels \( y_A \) and \( y_B \). The goal of CutMix is to create a new image \( \tilde{x} \) and label \( \tilde{y} \) by combining \( (x_A, y_A) \) and \( (x_B, y_B) \), as described below.

The combination operation is defined as:

\begin{equation}
\tilde{x} = M \odot x_A + (1 - M) \odot x_B
\label{eq:cutmix_image}
\end{equation}

\begin{equation}
\tilde{y} = \lambda y_A + (1 - \lambda) y_B
\label{eq:cutmix_label}
\end{equation}

where \( M \in \{0, 1\}^{W \times H} \) is a binary mask indicating where to drop out and fill in from the two images. \( M \) is 1 in the region taken from image \( x_A \) and 0 in the region taken from image \( x_B \). The symbol \( \odot \) represents element-wise multiplication. The ratio \( \lambda \) is sampled from a uniform distribution \( \mathcal{U}(0, 1) \). The generated image \( \tilde{x} \) is a mixture of \( x_A \) and \( x_B \), and the label \( \tilde{y} \) is a weighted combination of \( y_A \) and \( y_B \). The new training sample \( (\tilde{x}, \tilde{y}) \) is then used to train the model with its original loss function.
We use cut-mix in augmentation to get better generalization due to the imbalance of the classes.

\subsection{Model Architectures and Comparative Experiments}
\subsubsection{Basic Models}
\begin{figure}[htbp]
    \centering
    \includegraphics[width=0.9\linewidth]{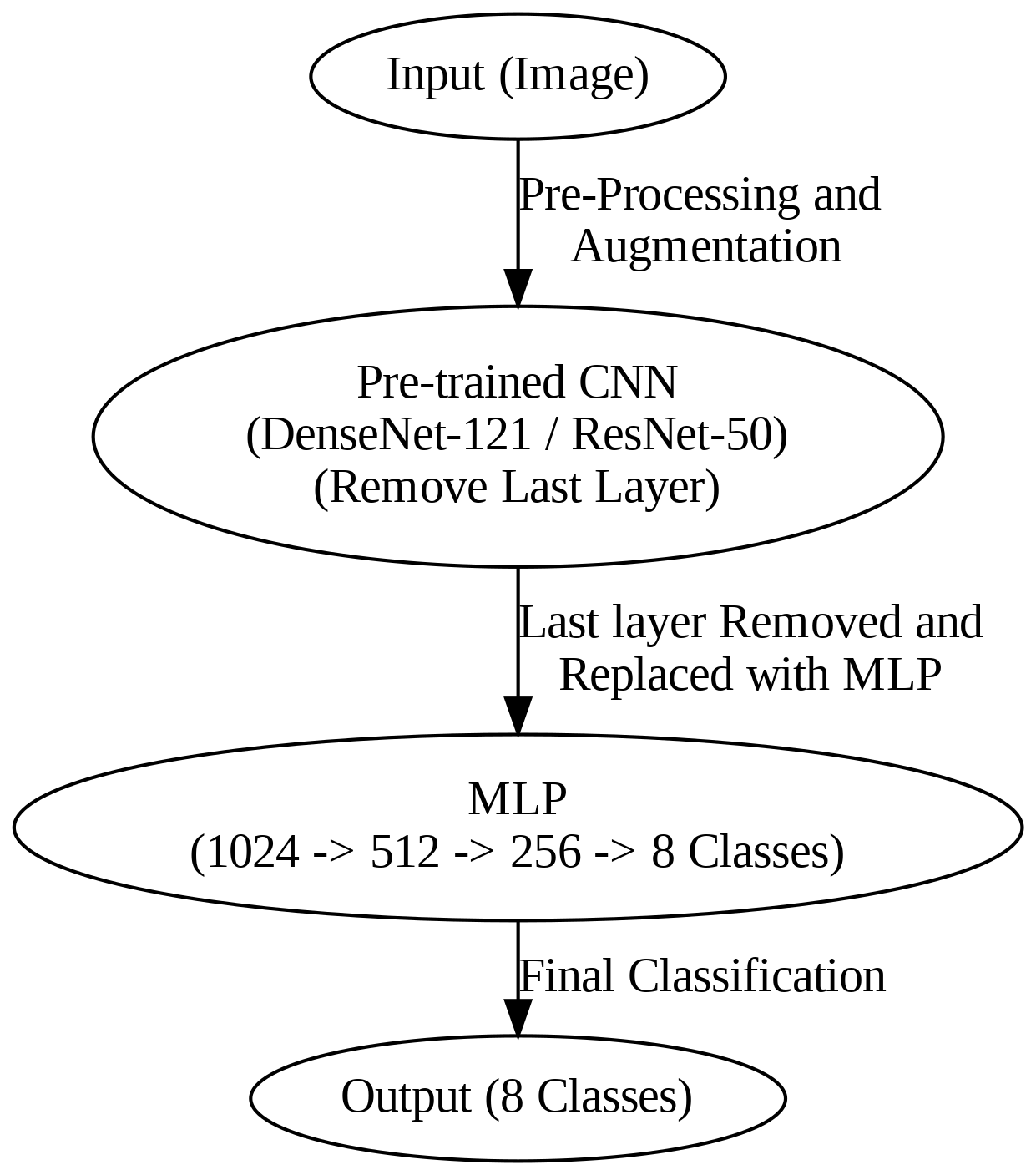}
    \caption{Architecture of Basic Model}
    \label{fig:model_arch}
\end{figure}

Our aim was to build a model which can also be incorporated in Internet of Things (IoT) devices. Hence, we selected lightweight models such as ResNet-50 \cite{resnet} and DenseNet-121\cite{densenet}, in order to balance performance and resource-constraints. ResNet-50 has 25.6 million parameters and DenseNet-121 has about 8 million parameters. The models are pre-trained on ImageNet Dataset \cite{deng2009imagenet}. The last layer is removed and is replaced by a Multi-Layer Perceptron as shown in Figure~\ref{fig:model_arch}. The model is then finetuned on the dataset.

\subsubsection{Models with CBAM}
\begin{figure}[htbp]
    \centering
    \includegraphics[width=0.9\linewidth]{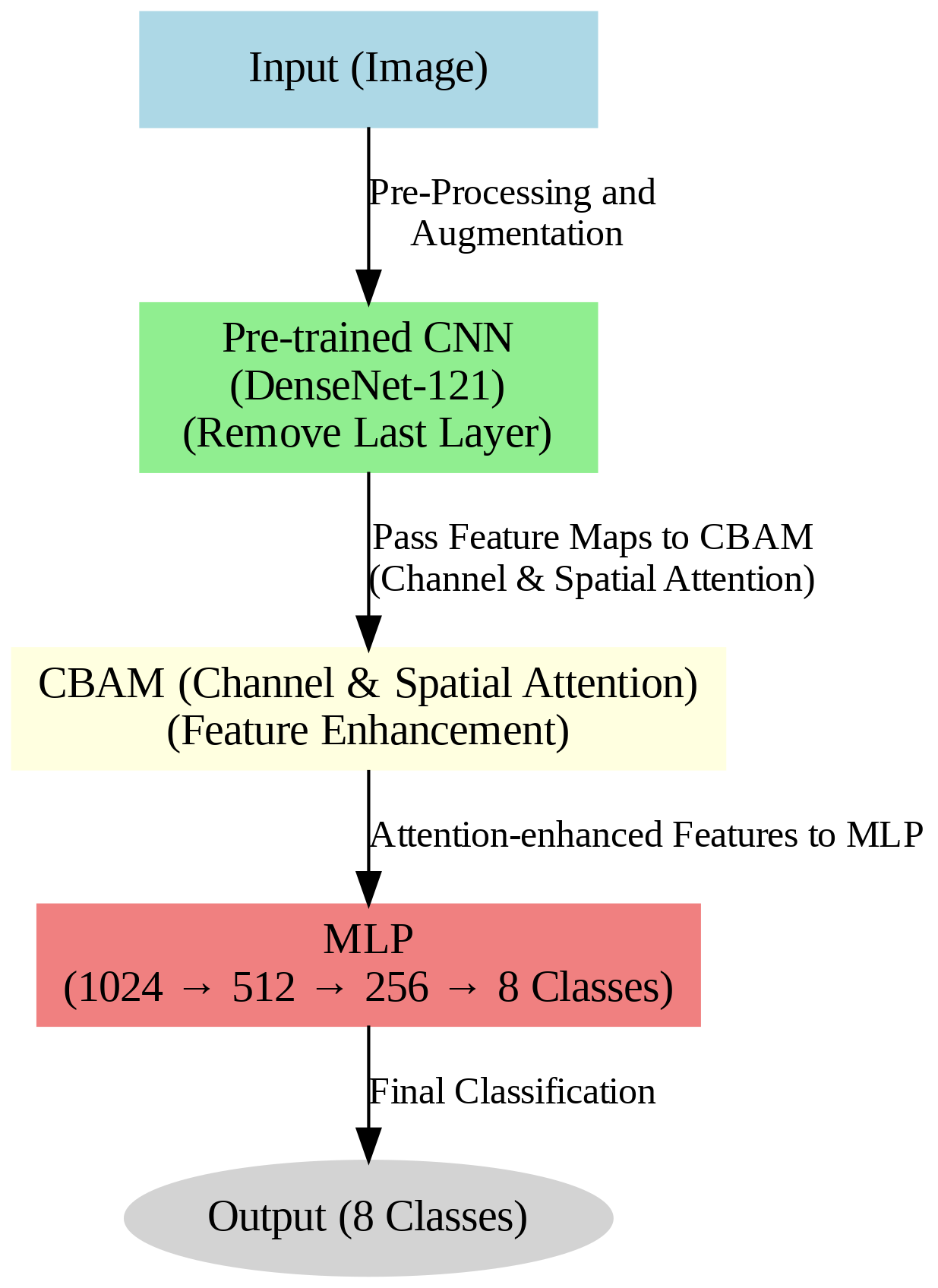}
    \caption{Architecture of Model with CBAM}
    \label{fig:model_arch_cbam}
\end{figure}

DenseNet-121 model was combined with Convolutional Block Attention Module (CBAM) \cite{woo2018cbam}. The CBAM module uses attention mechanism and enhances the feature representation by focusing on the informative parts of the feature representations. The Channel Attention Module (CAM) captures the important feature channels and the Spatial Attention Module (SAM) highlights important spatial regions. The architecture of the enhanced model using DenseNet-121 is shown in the Figure~\ref{fig:model_arch_cbam}. After the features are enhanced by CBAM, it is passed through a MLP and finally classified into the eight classes.

\subsection{Loss Functions and Training Procedure}
\subsubsection{Weighted Cross-Entropy Loss}
We used weighted-cross entropy loss (WCE) in order to mitigate the effects of weight imbalance in the dataset. The loss is as given in Equation~\ref{eq:wce}.

\begin{equation}
L_{\text{WCE}} = -\frac{1}{N} \sum_{i=1}^{N} w_{y_i} \log(p_{y_i})
\label{eq:wce}
\end{equation}

where:
\begin{itemize}
    \item \( N \) is the total number of samples in the batch.
    \item \( y_i \) is the true class label for sample \( i \).
    \item \( p_{y_i} \) is the predicted probability for the correct class \( y_i \).
    \item \( w_{y_i} \) is the weight associated with class \( y_i \)
\end{itemize}

Class weights $W_{y_i}$ are computed as given in equation~\ref{eq:weights}
\begin{equation}
w_{y_i} = \frac{\text{max\_count}}{\text{count}_{y_i}}
\label{eq:weights}
\end{equation}

where:
\begin{itemize}
    \item \( \text{max\_count} \) is the maximum number of images in any class.
    \item \( \text{count}_{y_i} \) is the number of images in class \( y_i \).
\end{itemize}

\subsubsection{Focal Loss}
Another method to mitigate the effects of class imbalance is by using Focal Loss \cite{focal_loss}. The loss is give as in Equation~\ref{eq:focal_loss}. It reduces the loss on easy or  majority class data and focuses more on hard or minority class data.

\begin{equation}
\mathcal{FL}(p_t) = -\alpha_t (1 - p_t)^\gamma \log(p_t)
\label{eq:focal_loss}
\end{equation}
where:
\begin{itemize}
    \item $p_t$ is the estimated probability for the true class,
    \item $\alpha_t$ are weights for different classes,
    \item $\gamma \geq 0$ is the focusing parameter which is tuned to reduce the loss for well-classified examples
\end{itemize}

\subsection{Evaluation Metrics}
After training the model, it is evaluated on the test data. These metrics are calculated on the test dataset. As our dataset is highly imbalanced, overall accuracy may be a misleading metric. Hence, we have used, class-wise accuracies and F1 score, Overall accuracy, F1 score,Precision, Recall and confusion matrix as our metrics. The main metric used for evaluation is the F1 score.
The confusion matrix is used to evaluate the following:-
\begin{itemize}
    \item \(TP_i\) (True Positives): Number of samples correctly predicted as class \(i\).
    \item \(FP_i\) (False Positives): Number of samples incorrectly predicted as class \(i\).
    \item \(FN_i\) (False Negatives): Number of samples of class \(i\) incorrectly predicted as other classes.
    \item \(N_i\): Total number of samples in class \(i\).
\end{itemize}

Using these, we define the evaluation metrics as follows:

\paragraph{Class-wise Accuracy:}
\begin{equation}
\text{Accuracy}_i = \frac{TP_i}{N_i}
\end{equation}

\paragraph{Overall Accuracy:}
\begin{equation}
\text{Accuracy} = \frac{\sum_{i=1}^C TP_i}{\sum_{i=1}^C N_i}
\end{equation}

\paragraph{Class-wise Precision and Recall:}
\begin{equation}
\text{Precision}_i = \frac{TP_i}{TP_i + FP_i}
\quad,\quad
\text{Recall}_i = \frac{TP_i}{TP_i + FN_i}
\end{equation}

\paragraph{Class-wise F1 Score:}
\begin{equation}
F1_i = 2 \times \frac{\text{Precision}_i \times \text{Recall}_i}{\text{Precision}_i + \text{Recall}_i}
\end{equation}

\paragraph{Overall F1 Score:}
\begin{equation}
F1_{overall} = \frac{1}{C} \sum_{i=1}^C F1_i
\end{equation}

\section{Experiments and Results}
We divide the dataset into training and test splits with 80\% training, 20\% test. We further divide the training set into 80\% train and 20\% validation sets. The experiments are conducted using five fold cross validation for hyper-parameter tuning and for checking the robustness of the models. 

The first experiment was to compare results obtained from ResNet-50 and DenseNet-121 models and compare which give better results, we also compare Pipleline B (Imbalanced Sampler) with Cross Entropy Loss (CE) with Pipeline A (Normal Augmentation) with Weighted Cross Entropy Loss.

\begin{table}[h]
\centering
\caption{Comparison of Imbalanced Sampler \& Weighted Cross Entropy Loss with ResNet-50 and DenseNet-121, \\for 9 classes (1 Healthy, 8 Disease or Pest). \\Values represent overall accuracy (\%).}
\label{tab:resnet_densenet}
\begin{tabular}{|>{\raggedright\arraybackslash}p{4.5cm}|c|c|}
\hline
\textbf{Training Method} & \textbf{ResNet-50} & \textbf{DenseNet-121} \\
\hline
Imbalanced Sampler (Pipeline B) with Cross Entropy Loss & 78.28 & \textbf{82.44} \\
\hline
Normal Augmentation (Pipeline A) with Weighted Cross Entropy Loss & 87.15 & \textbf{90.99} \\
\hline
\end{tabular}
\end{table}

From Table~\ref{tab:resnet_densenet}, we observe that DenseNet-121 performs better than ResNet-50 and also, Weighted Cross Entropy (WCE) loss, performs better than Imbalanced Dataset Sampler. Hence in further experiments, we use DenseNet-121 model. We also prefer using Unbalanced Dataset mitigating losses, instead of trying to make the dataset balanced.

\begin{figure}[htbp]
    \centering
    \includegraphics[width=0.9\linewidth]{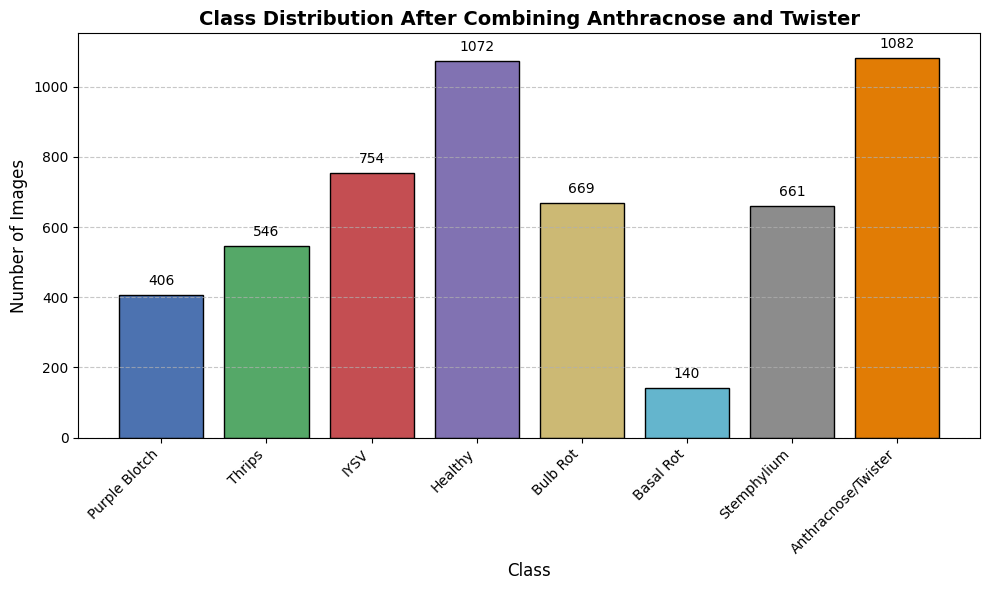}
    \caption{Dataset Distribution after combining Anthracnose and Twister classes}
    \label{fig:anthra_twist_comb}
\end{figure}

In the dataset, Anthracnose and Twister are labeled as separate classes; however, the National Library of Medicine states that Twister and Anthracnose are the same diseases \cite{anthra_twister}. Hence, for further experiments, we combine anthracnose and twister classes, to obtain one healthy class and seven diesease or pest classes. The dataset distribution after combining Anthracnose and Twister classes is as shown in Figure~\ref{fig:anthra_twist_comb}.

We then experiment on this new dataset, with Models as given below:-
\begin{itemize}
    \item Basic Model DenseNet-121 with Weighted Cross Entropy Loss and Pipeline A (Normal Augmentation), as shown in Figure~\ref{fig:model_arch}.
    \item DeseNet-121 model with CBAM, Weighted Cross Entropy Loss and Pipeline A (Normal Augmentation)
    \item Basic Model DenseNet-121, Weighted Cross Entropy Loss and Pipeline C (Albumentations-Based Augmentation)
    \item Basic Model DenseNet-121, Weighted Cross Entropy Loss and Pipline D (Cut-Mix Augmentation)
    \item DenseNet-121 model with CBAM, Weighted Cross Entropy Loss, Pipeline C (Albumentation-based Augmentaion) and Pipeline D (Cut-Mix Augmentation)
    \item DenseNet-121 model with CBAM, Weighted Cross Entropy Loss, Pipeline D (Cut-Mix Augmentation)
    \item DenseNet-121 model with CBAM, Focal Loss, Pipeline A (Normal Augmentation)
\end{itemize}

The comparative results of these Models is given in Table~\ref{tab:model_comparisons}.

\begin{table*}[htbp]
\centering
\renewcommand{\arraystretch}{1.2}
\begin{tabular}{|c
  |>{\centering\arraybackslash}p{1.9cm}
  |>{\centering\arraybackslash}p{1.9cm}
  |>{\centering\arraybackslash}p{1.9cm}
  |>{\centering\arraybackslash}p{1.9cm}
  |>{\centering\arraybackslash}p{1.9cm}
  |>{\centering\arraybackslash}p{1.9cm}
  |>{\centering\arraybackslash}p{1.9cm}|}
\hline
\textbf{Metric} & \textbf{D121 + WCE + A} & \textbf{D121 + CBAM + WCE + A} & \textbf{D121 + CBAM + Focal + A} & \textbf{D121 + WCE + C} & \textbf{D121 + WCE + D} & \textbf{D121 + CBAM + WCE + C+D} & \textbf{D121 + CBAM + WCE + D}  \\
\hline
Overall Accuracy (in \%) & 94.10 & 95.12 & 94.56 & 91.93 & 96.00 & 93.06  & 96.90 \\
\hline
Macro Precision & 0.90 & 0.93  & 0.93 & 0.93 & 0.94 & 0.91 & 0.96 \\
\hline
Macro Recall & 0.92 & 0.95 & 0.93 & 0.92 & 0.94 & 0.92 & 0.95 \\
\hline
Macro F1 Score & 0.91 & 0.94 & 0.93 & 0.92 & 0.94 & 0.92 & 0.96 \\
\hline
\end{tabular}
\smallskip
\caption{Comparison of DenseNet-121 configurations and training pipelines. Abbreviations: D121 = DenseNet-121, WCE = Weighted Cross Entropy, CBAM = Convolutional Block Attention Module, A = Pipeline A (Normal Augmentation), C = Pipeline C (Albumentations), D = Pipeline D (CutMix).}
\label{tab:model_comparisons}
\end{table*}

\begin{figure}[htbp]
    \centering
    \includegraphics[width=0.9\linewidth]{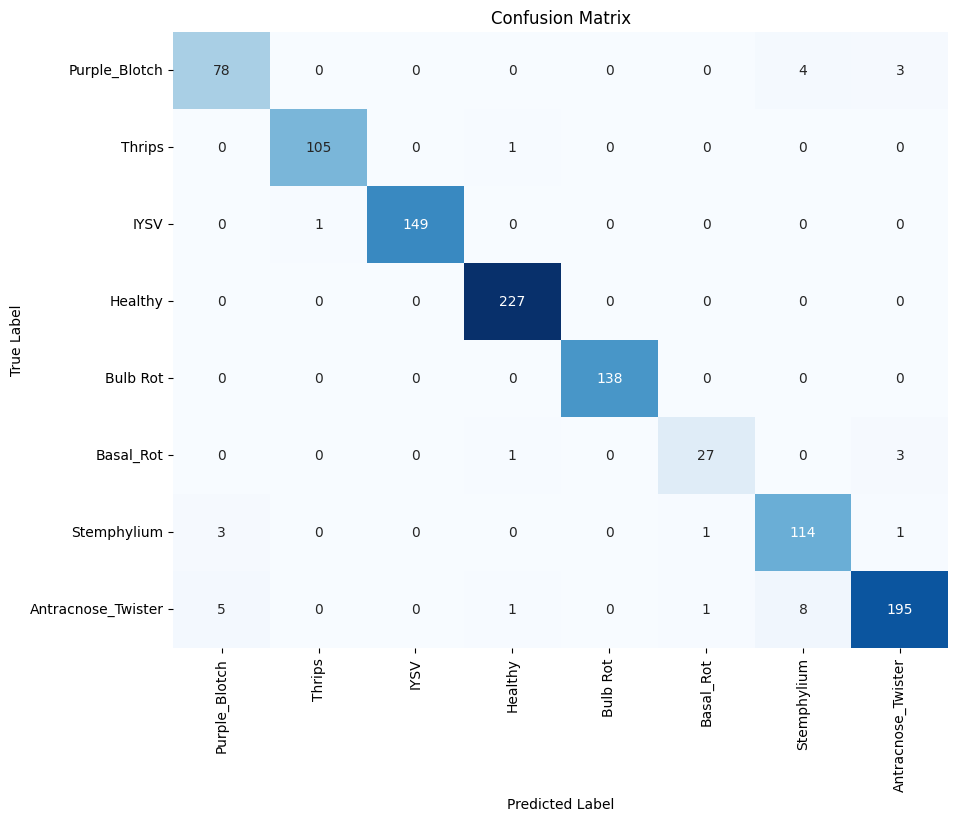}
    \caption{Confusion Matrix on test data by Densenet-121 with CBAM, WCE, Pipeline D (Cut-Mix)}
    \label{fig:confusion_matrix}
\end{figure}

From Table~\ref{tab:model_comparisons}, we observe that DenseNet-121 with CBAM, Weighted Cross Entropy Loss and Pipeline D (Cut-Mix) gives us the best results. The Confusion Matrix of the test data on this Model is shown in Figure~\ref{fig:confusion_matrix}

The class-wise accuracies obtained with this model is shown in Table~\ref{tab:class_wise_accuracies}.

\begin{table}[h]
\centering
\caption{Class-Wise Accuracies and F1 score obtained by Densenet-121 with CBAM, WCE, Pipeline D (Cut-Mix)}
\label{tab:class_wise_accuracies}
\begin{tabular}{|>{\raggedright\arraybackslash}p{2.5cm}|c|c|}
\hline
\textbf{Class-Name} & \textbf{Accuracy} & \textbf{F1-Score} \\
\hline
 Purple Blotch &0.91 & 0.91 \\
\hline
 Thrips & 0.99 & 0.99 \\
\hline
 IYSV & 0.99 & 0.99 \\
\hline
Healthy & 1.00 &  0.99\\
\hline
Bulb Rot & 1.00 &  1.00\\
\hline
Basal Rot & 0.87 & 0.91 \\
\hline
Stemphylium & 0.95 &  0.93\\
\hline
Anthracnose/Twister & 0.92 & 0.95 \\
\hline
\end{tabular}
\end{table}

\section{Discussion}

\begin{figure}[htbp]
    \centering
    \includegraphics[width=0.9\linewidth]{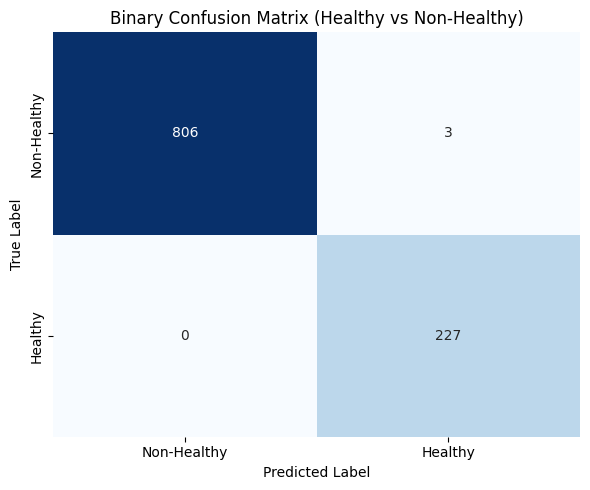}
    \caption{Confusion Matrix on test data by Densenet-121 with CBAM, WCE, Pipeline D (Cut-Mix) for Two Class}
    \label{fig:confusion_two_class}
\end{figure}

Our model mitigates the issue of imbalance in the classes. We obtain that all of the classes are being classified with a class accuracy greater than 80\%.
As an ablation study, we do an experiment to classify the imaged as healthy and un-healthy. We obtain accuracy as shown in Table~\ref{tab:two_class}. We observe that almost all the images are classified correctly. The Confusion matrix is as shown in Figure~\ref{fig:confusion_two_class}.

\begin{table}[h]
\centering
\caption{Two-Class, Healthy and Un-healthy classificatuion \\ Using DenseNet121 + CBAM + WCE + Cut-Mix}
\label{tab:two_class}
\begin{tabular}{|>{\raggedright\arraybackslash}p{2.5cm}|c|c|}
\hline
\textbf{Class} & \textbf{Accuracy} \\
\hline
Healthy &  1.00 \\
\hline
Un-healthy & 0.96 \\
\hline
\end{tabular}
\end{table}

We observe that as our model does not use localization of the region of disease as used in Raj et. al. \cite{raj2025}, Also, the number of images used for training is higher than 279 images per-class, we obtain better results compared to them. This is because our model does not have to focus on the bounding box, before classification, which is prone to errors due to annotation and lack of data for minority classes. Also, we have classified eight classes compared to five.

\section{Conclusion}
We address the issue of class imbalance in onion crop disease classification by exploring various image augmentation techniques and loss functions. Cut-Mix augmentation proved to be the most effective in mitigating the class imbalance. Weighted Cross-Entropy also outperforms Focal Loss and imbalanced dataset sampler. We further improved the model performance of DenseNet-121 model by incorporating Convolutional Blocal Attention Module (CBAM), which enhances the model's ability to focus on spatial and channel based on attention based features. Unlike, localization based approaches, our model does not depend on region specific annotations, reducing annotation based errors and complexity in model. 

The proposed model achieves 96.90\% accuracy and 0.96 F1 score on real-world onion disease field images, demonstrating the robustness of the model and it's practical applicability for effective pest management.



\end{document}